\documentclass[10pt,twocolumn,letterpaper]{article}

\usepackage{cvpr}              

\usepackage{graphicx}
\usepackage{amsmath}
\usepackage{amssymb}
\usepackage{booktabs}
\usepackage{float}
\usepackage[accsupp]{axessibility}

\usepackage[pagebackref,breaklinks,colorlinks]{hyperref}

\usepackage[capitalize]{cleveref}
\crefname{section}{Sec.}{Secs.}
\Crefname{section}{Section}{Sections}
\Crefname{table}{Table}{Tables}
\crefname{table}{Tab.}{Tabs.}

\def\cvprPaperID{8050} 
\def\confName{CVPR}
\def\confYear{2023}

\begin{document}

\title{Cross-Domain Image Captioning with Discriminative Finetuning}

\author{Roberto Dess\`i\\
Meta AI / UPF \\
\and
Michele Bevilacqua\\
Samaya AI\\
\and
Eleonora Gualdoni \\
UPF \\
\and
Nathanaël Carraz Rakotonirina \\
UPF \\
\and
Francesca Franzon \\
UPF \\
\and
Marco Baroni \\
UPF / ICREA \\
}
\maketitle

\begin{abstract}
Neural captioners are typically trained to mimic human-generated references without optimizing for any specific communication goal, leading to problems such as the generation of vague captions. In this paper, we show that fine-tuning an out-of-the-box neural captioner with a self-supervised discriminative communication objective helps to recover a plain, visually descriptive language that is more informative about image contents. Given a target image, the system must learn to produce a description that enables an out-of-the-box text-conditioned image retriever to identify such image among a set of candidates. We experiment with the popular ClipCap captioner, also replicating the main results with BLIP. In terms of similarity to ground-truth human descriptions, the captions emerging from discriminative finetuning lag slightly behind those generated by the non-finetuned model, when the latter is trained and tested on the same caption dataset. However, when the model is used without further tuning to generate captions for out-of-domain datasets, our discriminatively-finetuned captioner generates descriptions that resemble human references more than those produced by the same  captioner without finetuning. We further show that, on the Conceptual Captions dataset, discriminatively finetuned captions are more helpful than either vanilla ClipCap captions or ground-truth captions for human annotators tasked with an image discrimination task.\footnote{Our code is available at \url{https://github.com/facebookresearch/EGG/tree/main/egg/zoo/discriminative_captioner}.}
\end{abstract}


\begin{figure}[tb]
    \centering
    \includegraphics[width = \hsize]{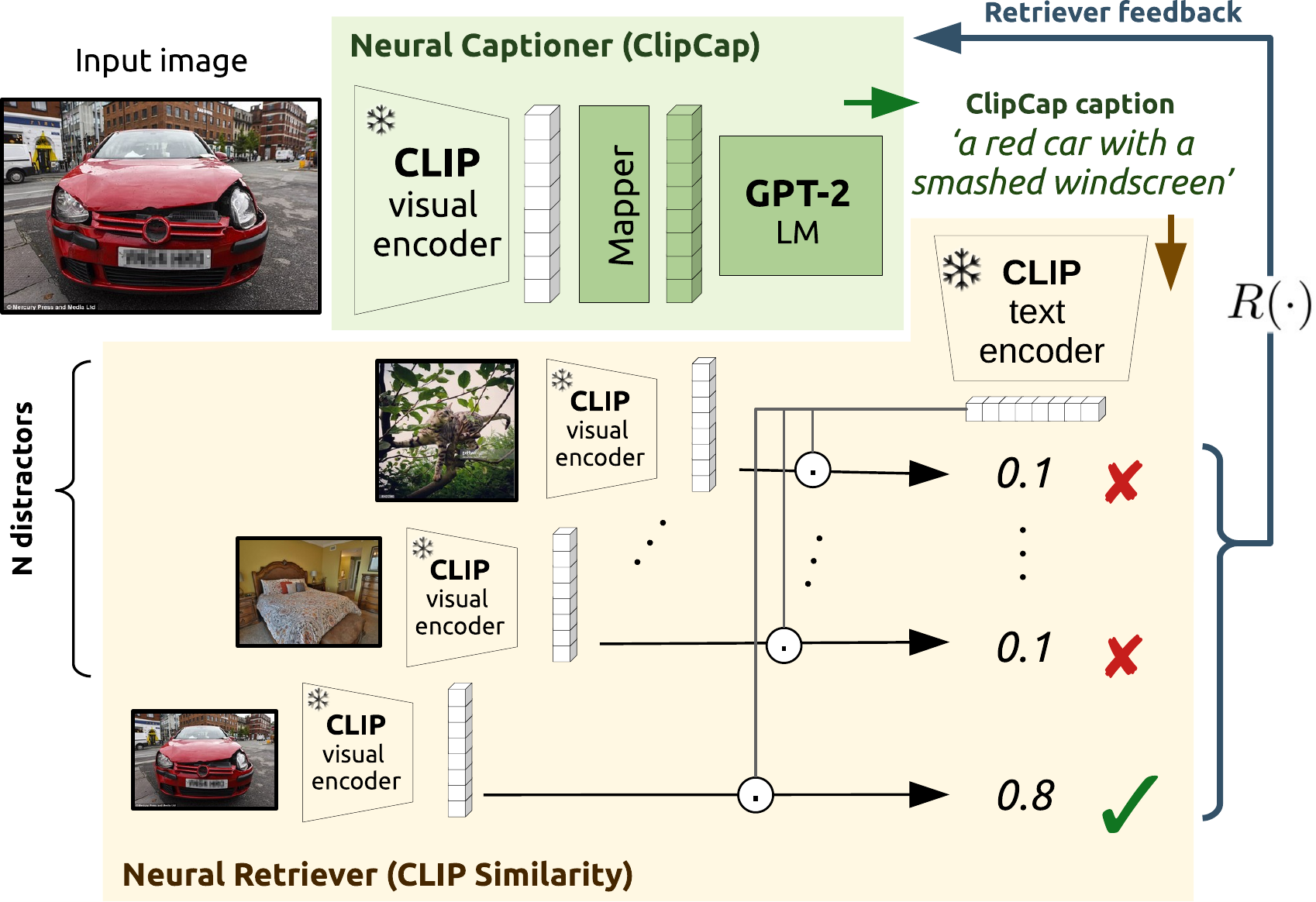}
    \caption{Setup of our discriminative finetuning method when applied to the ClipCap captioner~\cite{Mokady:etal:2021}. All CLIP encoders are frozen, while the language generation modules (mapper and GPT-2) are updated based on reward values. 
    }
    \label{fig:setup}
\end{figure}

\section{Introduction}

The last decade has seen impressive progress on the task of automatically generating image descriptions with deep neural networks \cite{Vinayls:etal:2015,Xu:etal:2015,Bernardi:etal:2016,Stefanini:etal:2022}. Most of the proposed methods try to optimize the similarity of system-produced captions with ground-truth human references, either through a standard cross-entropy cost function~\cite{Anderson:etal:2018}, or by maximizing natural-language-generation (NLG) metrics such as CIDEr \cite{Sai:etal:2022, Pinto:etal:2023} through a reward-based objective. While imitating human captions is a reasonable goal, it does not take into account that, in concrete applications, an image description is produced for a purpose \cite{Fisch:etal:2020,Kreiss:etal:2022}.

There are a multitude of context-dependent purposes a description might be produced for, but a fundamental one is to correctly characterize an object so that a hearer could \textit{discriminate} it from other contextual elements \cite{Kazemzadeh:etal:2014}. This ability to discriminate between referents is a core purpose of communication, playing a fundamental role in its evolution and acquisition (e.g., \cite{Skyrms:2010, Bloom:2000}). We study here what happens when we take an out-of-the-box image captioner that was trained to imitate human captions, and finetune its language components with a \textit{discriminative} objective using reinforcement learning. In particular, we let the captioner play a discrimination game with an out-of-the-box caption-based image retriever. The captioner generates a caption given a target image, and the retriever (whose weights are not updated) uses the caption to select the target among a set of candidates, as shown in Fig.~\ref{fig:setup}.  This finetuning technique does not require annotated data (only a set of images), and it's agnostic to the underlying captioner and retriever components.

We report two strong and novel results. First, we show that captions finetuned in this way lead to better 0-shot cross-domain caption generation.\footnote{We note that in the earlier literature, cross-domain captioning can rely on sets of unpaired images and captions from the target domain (e.g., \cite{Chen:etal:2017,Yang:etal:2017,Zhao:etal:2021}). We consider here a more challenging and realistic 0-shot transfer task in which \textit{no} captions from the target domain are available.} Second, not only are the finetuned captions good for neural text-based image retrieval (both in- and across-domain), but they can also be more useful to human annotators, helping discriminate the target from distractors \textit{more than human-generated ground-truth captions do}. We conclude the paper with an analysis of the finetuned captions, comparing them to human-generated and non-finetuned ones from the Conceptual Captions dataset.  
We find that discriminative finetuning undoes the more abstract language that the underlying system had learned from the ground-truth captions, leading to a more plainly descriptive style that we expect to be more useful in practical applications.

\section{Related Work}
\label{sec:related-work}

In recent years, deep learning has led most progress in image captioning \cite{Anderson:etal:2018, Donahue:etal:2017}. 
While early approaches relied on supervised learning to maximize the likelihood of the model against human references\cite{Anderson:etal:2018, Vinayls:etal:2015}, other methods have tried to maximize a reward based on standard language generation metrics \cite{Rennie:etal:2017, Ranzato:etal:2016}.  Our work belongs to a third tradition, exploring the idea of caption learning or finetuning with a reward-based objective that is not (only) based on a comparison with reference captions. Among the approaches more closely related to ours, the system of Yu \textit{et al.~}\cite{Yu:etal:2022} uses a ClipCap-like system for caption generation, and CLIP to measure image-caption similarity, focusing on generating captions in multiple styles. Cho \textit{et al.~}\cite{Cho:etal:2022} use CLIP to finetune a pre-trained captioner. Like \cite{Yu:etal:2022}, they use the CLIPScore image-caption similarity measure \cite{Hessel:etal:2021} as a reward signal, rather than a discriminative objective like the one we adopt. We show in Appendix B.2 
that the discriminative objective outperforms an image-caption similarity objective similar to CLIPScore (\textit{cosine similarity} in Table 9).  
Luo \textit{et al.~}\cite{Luo:etal:2018} apply discriminative finetuning to basic MLE-trained models. Their method requires the use of ground-truth captions for concurrent CIDEr-based optimization.
A similar approach was presented by Liu \textit{et al.~}\cite{Liu:etal:2018}. They also finetune a MLE-pre-trained captioner with a mixture of discriminative and CIDEr-based rewards (requiring ground-truth captions for the latter).
Intriguingly, their method produces a higher portion of novel and unique captions compared to a system trained only to imitate human captions, pointing to the potential for generalization that we confirm here. Finally, in another early paper using discrimination-based caption learning, Dai and Lin \cite{Dai:Lin:2017} propose a loss that explicitly pushes a model to make captions more discriminative with respect to those of a reference model. 

Our study builds on this earlier work. Our main novelties lie in the development of a simple system to perform discriminative finetuning that only requires unannotated images, out-of-the-box generation/retrieval components and applying the vanilla REINFORCE algorithm, and in our out-of-domain and human-based evaluations.

\section{Discriminative Self-Supervised Training} 
\label{sec:experimental-setup}
We finetune with reinforcement learning a pre-trained image captioner to generate a caption that is subsequently fed as input to a frozen discriminator. The discriminator is then tasked to retrieve the original image among a set of distractors. We refer to our tuning method as \textit{DiscriTune}.

\paragraph{Captioner} We employ a multimodal language model as neural captioner. 
We experiment with two different captioners, namely the ClipCap architecture from Mokady \etal~\cite{Mokady:etal:2021}, and the captioner network of the BLIP model~\cite{Li:etal:2022}.\footnote{Additional results with the recently introduced CaMEL~\cite{Barraco:etal:2022} captioner are provided in Appendix C.}

ClipCap is an image-conditioned publicly available GPT-2 model \cite{Radford:etal:2019}. 
Importantly, Mokady \etal~\cite{Mokady:etal:2021} showed that it performs comparably to other state-of-the-art captioning systems while having less parameters and being more efficient to train.
Given an input image, ClipCap uses a frozen CLIP \cite{Radford:etal:2021} visual encoder to extract visual features. These features are then projected through a trainable mapper network onto the GPT-2 embedding space, where they are used as a prefix conditioning the generation of an image description. In practice, they act as soft image prompts that are used to kickstart the caption generation process.

We use the publicly available ClipCap checkpoints where the language model weights were also updated during training. Such checkpoints were trained on Conceptual Captions \cite{Sharma:etal:2018} (\textit{ClipCap-ConCap} henceforth), and MS COCO \cite{Lin:etal:2014} (\textit{ClipCap-COCO} henceforth), respectively. Given that these models were trained with a frozen  ViT-B/32-based CLIP as visual feature extractor, we use the same visual encoder in our experiments.

To test whether our method generalizes to other models, we experiment with the BLIP system~\cite{Li:etal:2022}, a large captioner pre-trained with web-mined data which performed on par or better than other state-of-the-art systems on several benchmarks and has shown strong retrieval and captioning performance \cite{Li:etal:2022}. BLIP is made of a text Transformer~\cite{Vaswani:etal:2017} and a vision Transformer~\cite{Dosovitskiy:etal:2021}. The text Transformer is trained to maximize the likelihood of reproducing a ground-truth reference by autoregressively generating a caption. Visual information is injected by cross-attending over the output of the vision Transformer. The model is trained with multi-task learning with two other multimodal alignment losses.
We refer the reader to~\cite{Li:etal:2022} for additional details on BLIP. We use the BLIP-base version,\footnote{The model is publicly available through the LAVIS library \cite{Li:etal:2022lavis}} pre-trained on a large dataset including both COCO and Conceptual Captions. 

\paragraph{Retriever} We used the standard CLIP model from Radford \etal~\cite{Radford:etal:2021} as our neural retriever. CLIP is a multimodal dual encoder model that embeds text and images and learns to maximize their similarity through a contrastive loss. It was trained on a dataset of 400M human-curated image-text pairs. We refer the reader to \cite{Radford:etal:2021} for additional details about CLIP pre-training. We use the publicly available original implementation of CLIP with a ViT-B/32  backbone \cite{Dosovitskiy:etal:2021}.~\footnote{\url{https://github.com/openai/CLIP}}

We perform retrieval by computing a matching score $\mathrm{match}(c, i)$ for an image $i$ and a caption $c$ as the dot product between the embedded representation of $c$ (computed by the frozen CLIP textual encoder) and the embedded representation of $i$ (computed by the frozen CLIP visual encoder). We consider an image $i$ correctly retrieved against a set of distractors $D \subseteq I$ (taken from a larger image collection $I$) iff $i' \in D$, and $i \neq i'$, $\mathrm{match}(c, i) > \mathrm{match}(c, i')$.

\paragraph{Optimization}
Since the decoding process creates a discrete bottleneck, we cannot use a loss function to backpropagate end-to-end from the retriever output. Thus, we use REINFORCE \cite{Williams:1992} to optimize the captioner (ClipCap or BLIP) by using a reward that is based on the matching scores between the caption $c$ and the set of candidates $D \cup \{i\}$. In this process, the CLIP retriever is kept frozen and, given the non-differentiable text generation step, there is no gradient flowing from the retriever to the captioner. 

To compute a reward, we normalize the retriever scores to a probability distribution, and then use as reward the so-defined log probability of the original image $i$:
$
    R(c, i, D) = \mathrm{log} \ \frac{
    e^{\mathrm{match}(c, i)}}{
    \sum_{i' \in D \cup \{i\}}e^{\mathrm{match}(c, i')}
    }
$. We then compute the reward as the cross-entropy loss between such distribution and the position of the target image in the list of candidates.
Our captioner, which, in essence, models the probability $P_{\theta}(c|i)$ of a caption $c$ given an image $i$, is, in reinforcement learning terms, the action policy, and the actions taken are just the selected tokens. The policy is trained to minimize the negative expected reward, i.e., $\mathbb{E}_{c \sim P_{\theta}(\cdot|i)}[-R(c, i, D)]$ and we compute the gradient of this expectation as the expectation of the gradient.
To reduce variance of the gradient estimator we use as a baseline a running mean of past rewards \cite{Sutton:Barto:2018} (see Appendix B for more details).

In our experiments we let both our captioners freely generate for a maximum of 40 tokens, or until a full stop (or EOS token for BLIP) is produced. We set such maximum length after observing that most captions were much shorter and taking into account the fact that CLIP, our downstream retriever, does not process contexts larger than 75 tokens.

\section{Experiments}
\label{sec:experiments}

\subsection{Setup}

In this section, we report experiments with our DiscriTune method where we finetune the pre-trained captioner with a self-supervised discriminative reward provided by a frozen downstream CLIP receiver.
We apply the method to the ClipCap checkpoints introduced above, namely  ClipCap-COCO and ClipCap-ConCap. We finetune them with CLIP-provided discriminative reward using COCO and Conceptual Captions data, respectively. We call the resulting finetuned models DiscriTune-COCO and DiscriTune-ConCap. We also repeat the main experiments using the BLIP checkpoint trained on COCO, which we further finetune on COCO with our DiscriTune method.\footnote{We use the EGG library~\cite{Kharitonov:etal:2021} to perform all our experiments.} 

We finetune both our captioners with the discriminative reward using the Adam \cite{Kingma:Ba:2014} optimizer with a learning rate of $10^{-7}$ and a constant schedule.
We use a batch size of 100, with in-batch distractors, so that each target image is mixed with 99 randomly sampled distractors for discriminative finetuning. All our experiments were performed on a single Tesla V100 GPU.
For ClipCap on COCO, we finetune for $20$ epochs, with around 1.2K batches per epochs and roughly 24K updates, whereas on Conceptual Captions we finetune for $2$ epochs leading to around 28K batches per epoch and 56K updates.
For BLIP, we finetune on COCO for a single epoch corresponding to 1.2K updates, as preliminary experiments showed that to be sufficient for convergence.
We use greedy decoding at training time and beam search at test time.

\paragraph{Hyperparameter search} We used the Flickr validation set  \cite{Young:etal:2014}, not used elsewhere in the paper, to tune learning rate, reward function (discriminative- vs accuracy-based), and REINFORCE baseline (see Appendix B).

\paragraph{Data} We used the standard \textbf{MS COCO} \cite{Lin:etal:2014} and \textbf{Conceptual Captions} \cite{Sharma:etal:2018} datasets to perform discriminative finetuning. MS COCO is on of the most commonly used captioning and text-conditioned retrieval datasets, containing around 120K images, each provided with 5 human-generated captions.
We use the Karpathy train and test set split \cite{Karpathy:FeiFei:2014}. Conceptual Captions is a collection of images mined from the web and aligned with their alt-text descriptions. It contains around 3M samples for training and around 16K images for validation, that we used as our test set. After filtering out corrupted images due to outdated download links, we are left with 2.8M samples in the train set and exactly 13K images for testing.

To test 0-shot cross-domain generalization, we use the \textbf{Flickr} \cite{Young:etal:2014}, \textbf{nocaps} \cite{Agrawal:etal:2019} and \textbf{Concadia} \cite{Kreiss:etal:2022} datasets. 
We divided Flickr data according to the Karpathy split and evaluate on the test set (1K samples), where each image is aligned with 5 human captions. The nocaps dataset was introduced to test caption generalization performance of models trained on COCO. It is divided into three splits. We do not use the control \textit{in-domain} split, since it contains the same object classes as COCO. The \textit{near-domain} split has images from COCO categories as well as images from new categories. The \textit{out-domain} split only contains pictures of objects that are not present in COCO.  Following prior work \cite{Mokady:etal:2021}, we use the validation set of nocaps for testing purposes. Concadia is a dataset recently introduced to test difference in text generation performance when producing captions compared to descriptions. The former are meant to accompany an image in order to provide additional context, as in books or newspapers, while the latter should be able to replace the image, an example being descriptions for visually impaired people.\footnote{Note that in this paper we follow instead the standard practice of using ``caption'' to refer to both captions and descriptions in the sense of Kreiss \etal~\cite{Kreiss:etal:2022}.}
 We use the Concadia test split, which contains 9.6K images, each annotated with a caption and a description.

We want to emphasize again that, when finetuning our models, we did not use any human reference. Ground-truth captions were only used to compute NLG metrics.

\begin{table*}[tb]
\centering
\begin{tabular}{l|r|r|r|r|r|r|r}
Model & {\textit{COCO}} & {\textit{ConCap}} & \textit{Flickr} & \textit{nocaps near} & \textit{nocaps out} & \textit{Concadia} \\ 
\hline
ClipCap-COCO & 74.2 & 73.0 & 65.9 & 77.3& 73.9 & 53.74 \\
DiscriTune-COCO & \textbf{84.8} & 83.6 & 79.4& 86.0&  82.5 & 64.79 \\
\hline
ClipCap-ConCap& 73.4& 82.5 & 76.7& 78.1& 73.6 & 59.17 \\
DiscriTune-ConCap & 81.6& \textbf{94.4} & 87.8 & \textbf{89.1}& \textbf{88.7}& \textbf{80.49} \\
\hline
Human captions & 76.3 & 81.6 & \textbf{88.7} & 85.5 & 87.7& 73.96 \\
\hline\hline
\end{tabular}
\caption{ClipCap and DiscriTune percentage accuracy (P@1) when retrieving a target image from a set of 100 candidates taken from the COCO, Concpetual Captions and Concadida test set, and nocaps validation sets.}
\label{tab:retrieval-results}
\end{table*}

\subsection{Results}
\subsubsection{Text-Conditioned Image Retrieval}
\label{sec:image-retrieval}

Table~\ref{tab:retrieval-results} shows that, as expected, ClipCap trained with DiscriTune greatly improves over vanilla ClipCap on the image retrieval tasks it was finetuned on. The result however also extends to cross-domain retrieval. DiscriTune-COCO (slightly) outperforms ClipCap-ConCap on Conceptual Captions, and DiscriTune-ConCap greatly outperforms ClipCap-COCO on COCO. Moreover, both versions of DiscriTune greatly outperform ClipCap on all other datasets. 
Interestingly, DiscriTune almost always outperforms human-generated captions, confirming a recent result by Dess\`i \etal~\cite{Dessi:etal:2022} on how neural retrievers show better performance with neural captions. We performed additional experiments on a more challenging setup where hard distractors are either automatically mined or selected from adjacent frames in videos and confirm the superior performance of our DiscriTune method (Appendix D).

\begin{table}[tb]
    \centering
    \begin{tabular}{l|cccc}
    \multicolumn{5}{c}{\textit{COCO}}\\ 
    \hline
    Model& B@4& M& C& S \\
    ClipCap-COCO&  \textbf{32.60}& \textbf{27.50}& \textbf{108.55}& \textbf{20.33}\\
    DiscriTune-COCO& 32.31& 26.05& 105.40& 20.03 \\
    \hline\hline
    \multicolumn{5}{c}{\textit{Conceptual Captions}} \\
    \hline
    Model& B@4& M& C& S \\
    ClipCap-ConCap& \textbf{7.32}& \textbf{10.81}& \textbf{87.22}& \textbf{18.07} \\
    DiscriTune-ConCap& 3.92& 8.79& 55.26& 15.40 \\
    \hline\hline
    \end{tabular}
    \caption{NLG metrics (BLEU@4\cite{Papineni:etal:2002}, METEOR\cite{Denkowski:etal:2014}, CIDEr\cite{Vedantam:etal:2015} and SPICE\cite{Anderson:etal:2016}) for ClipCap and DiscriTune captions tested in-domain on COCO and Conceptual Captions.}
    \label{tab:first-results-nlg}
\end{table}

\begin{table}[tb]
    \centering
    \begin{tabular}{l|cccc}
    \multicolumn{5}{c}{\textit{COCO}}\\ 
    \hline
    Model& B@4& M& C& S \\
    ClipCap-ConCap&  8.50& 13.29& 37.03& 9.77 \\
    DiscriTune-ConCap&  \textbf{13.99}& \textbf{16.97}& \textbf{53.20}& \textbf{12.07} \\
    \hline\hline
    \multicolumn{5}{c}{\textit{Conceptual Captions}}\\ 
    \hline
    ClipCap-COCO&  1.47& 6.43& 23.74 & 7.98 \\
    DiscriTune-COCO& \textbf{1.71}& \textbf{6.58}& \textbf{28.01}& \textbf{9.00} \\
    \hline\hline
    \multicolumn{5}{c}{\textit{Flickr}}\\ 
    \hline
    ClipCap-COCO& 17.21& 18.43& 41.65& 12.04\\
    DiscriTune-COCO& \textbf{18.48}& \textbf{18.61}& \textbf{44.78}& \textbf{12.68} \\
    \hline
    ClipCap-ConCap& 8.28& 12.24& 27.57& 7.81 \\
    DiscriTune-ConCap& 13.01& 15.16& 36.35& 9.44 \\
    \hline\hline
    \multicolumn{5}{c}{\textit{nocaps-near}} \\
    \hline
    ClipCap-COCO& 30.47& \textbf{24.36} & 69.66 & 10.89 \\
    DiscriTune-COCO& \textbf{32.87}& 24.11& \textbf{70.63}& \textbf{10.98} \\
    \hline
    ClipCap-ConCap& 10.41& 13.25& 30.47& 5.72 \\
    DiscriTune-ConCap& 18.93& 17.66& 46.45& 7.53 \\
    \hline\hline
    \multicolumn{5}{c}{\textit{nocaps-out}} \\
    \hline
    ClipCap-COCO& 20.32& 20.22& 51.74 & 8.55 \\
    DiscriTune-COCO& \textbf{24.10}& \textbf{20.49}& \textbf{57.06}& \textbf{8.83} \\
    \hline
    ClipCap-ConCap& 10.69& 13.15& 36.57& 5.71 \\
    DiscriTune-ConCap& 16.76& 17.03& 54.03& 7.56 \\
    \hline\hline
    \multicolumn{5}{c}{\textit{Concadia-Descriptions}} \\
    \hline
    ClipCap-COCO& 1.94 & 5.57 & 14.99 & 6.44\\
    DiscriTune-COCO& \textbf{2.03} & \textbf{5.65} & 16.70 & \textbf{7.15}\\
    \hline
    ClipCap-ConCap& 0.62& 3.69& 12.77& 5.82\\
    DiscriTune-ConCap& 1.12& 4.81& \textbf{17.20}& 7.13\\
    \hline\hline
    \multicolumn{5}{c}{\textit{Concadia-Captions}} \\
    \hline
    ClipCap-COCO& 0.24& 2.45& 4.35 & 2.11 \\
    DiscriTune-COCO& 0.30 & 2.49 & 5.57 & 2.59\\
    \hline
    ClipCap-ConCap& \textbf{0.50} & 2.74& 9.22 & 3.65\\
    DiscriTune-ConCap& 0.30& \textbf{2.79}& \textbf{9.37}& \textbf{4.20} \\
    \hline\hline
    \end{tabular}
    \caption{NLG metrics (BLEU@4, METEOR, CIDEr and SPICE) for ClipCap and DiscriTune captions tested across domain on COCO (Conceptual Captions-based models only), Conceptual\-Captions (COCO-based models only), Flickr, nocaps, Concadia.}
    \label{tab:gen-results-nlg}
\end{table}

\subsubsection{Ground-Truth-Based Caption Quality Evaluation}
The text-based caption retrieval results show that our approach is remarkably good at the task it was finetuned on, also in the 0-shot cross-domain setup. However, it is less clear that retrieval-based finetuning should improve the captions' faithfulness to human ground-truth image descriptions. Table~\ref{tab:first-results-nlg} shows that, indeed, discriminative finetuning leads to some decrease in caption faithfulness, when testing on the dataset used for supervised pre-training. However, this performance drop (which, in the case of COCO is very small) is balanced by greater generalization performance, as shown in Table~\ref{tab:gen-results-nlg}. DiscriTune-COCO is the best domain transfer model across-the-board. In addition, DiscriTune-ConCap is also consistently outperforming its vanilla counterpart, ClipCap-ConCap, in this 0-shot cross-domain generalization setup.
Interestingly, for the Concadia dataset we have a (slightly) higher gain over plain ClipCap with the descriptions split rather than with the captions one, especially for DiscriTune-ConCap. Human references in the descriptions split were generated with the goal of replacing an image by describing its contents. This confirms that DiscriTune contributes to captions that are more suited for the ``communicative'' purpose of characterizing the crucial aspects of an image contents. Overall, the results suggest that, on the one hand, discriminative finetuning leads to captions that drift apart somewhat from the human descriptions that the model learnt to mimic. 
At the same time, though, this might allow the model to get away from the idiosyncrasies of a specific captioning style, leading to captions that better generalize to a wider set of unseen images and domains. 
As the experiment in Section \ref{sec:human-retrieval}  below shows, sometimes this drifting might be beneficial even when captioning images in the same domain.

\subsubsection{Applying Discriminative Finetuning to BLIP}

To verify the robustness of DiscriTune finetuning, we next apply it to BLIP. We finetune BLIP on COCO captions, one of the datasets it was pre-trained on. Tables \ref{tab:blip-retrieval-results}
(retrieval) and \ref{tab:blip-gen-results-nlg} (captioning) show that Discrirune-BLIP outperforms its non-finetuned counterpart for both retrieval performance and caption quality.
The results confirm that our captioner-agnostic discriminative finetuning is helpful even when applied to this latest-generation general-purpose vision-and-language model.
DiscriTune-BLIP outperforms its vanilla counterpart on all out-of-domain datasets tested.
Again, we note the greater performance boost on the Concadia Descriptions split compared to the Captions one, confirming the results obtained with ClipCap and the benefits of our finetuning method to generate captions that are closer to human references produced with the communicative intent of replacing an image.
Given that BLIP is not based on a CLIP encoder, this experiment also refutes the hypothesis that DiscriTune performance gains are simply due to using CLIP both as visual encoder for the captioner and as retriever, confirming the wider applicability of our method.

\begin{table}[tb]
\centering
\begin{tabular}{l|r|r|r|r}
Model & \textit{Flickr} &  \textit{nocaps} & \textit{nocaps} &  \textit{Concadia} \\ 
& & \textit{near} & \textit{out} &  \\
\hline
BLIP &75.1 &  87.0& 90.6 &  71.0 \\ 
DiscriTune-BLIP & \textbf{80.8} & \textbf{90.2}&  \textbf{92.6}& \textbf{74.9} \\
\hline\hline
\end{tabular}
\caption{BLIP and DiscriTune-BLIP percentage accuracy on the cross-domain retrieval task described in Section \ref{sec:image-retrieval}.}
\label{tab:blip-retrieval-results}
\end{table}

\begin{table}[tb]
    \centering
    \begin{tabular}{l|cccc}
    \multicolumn{5}{c}{\textit{Flickr}}\\ 
    Model& B@4& M& C& S \\
    \hline
    BLIP& 27.18& 22.74& 70.63 & 16.03 \\
    DiscriTune-BLIP& \textbf{28.19}& \textbf{23.61}& \textbf{74.77}& \textbf{16.9} \\
    \hline\hline
    \multicolumn{5}{c}{\textit{nocaps-near}} \\
    \hline
    BLIP& 44.91& 29.52& 107.13& 14.52 \\
    DiscriTune-BLIP& \textbf{45.46}& \textbf{29.81}& \textbf{108.70}& \textbf{14.97} \\
    \hline\hline
    \multicolumn{5}{c}{\textit{nocaps-out}} \\
    \hline
    BLIP&  \textbf{38.42}&	26.95& 105.63&	13.77\\
    DiscriTune-BLIP& 37.34&	\textbf{27.13}& \textbf{106.23}&	\textbf{14.14}\\
    \hline\hline
    \multicolumn{5}{c}{\textit{Concadia-Descriptions}} \\ 
    BLIP &  2.71& 6.90 & 26.02 & 9.75 \\ 
    DiscriTune-BLIP& \textbf{2.90} & \textbf{7.18} & \textbf{27.20} & \textbf{9.98}\\
    \hline\hline
    \multicolumn{5}{c}{\textit{Concadia-Captions}} \\
    BLIP &0.66 & 3.14& 9.97& 3.84 \\ 
    DiscriTune-BLIP& \textbf{0.74} & \textbf{3.35} & \textbf{10.84}& \textbf{3.96}\\
    \hline\hline
    \end{tabular}
    \caption{NLG metrics (BLEU@4, METEOR, CIDEr and SPICE) for BLIP and DiscriTune-BLIP captions tested across domains.}
    \label{tab:blip-gen-results-nlg}
\end{table}

\subsubsection{Human Text-Based Image Retrieval}
\label{sec:human-retrieval}

The results in tables \ref{tab:retrieval-results} and \ref{tab:first-results-nlg} show that, when testing on Conceptual Captions, DiscriTune-ConCap produces outputs that are less similar to human captions than those of ClipCap-ConCap, but it greatly outperforms the latter in text-based image retrieval accuracy. Recall that the ground-truth captions in Conceptual Captions come from the alt-texts associated to images harvested from the Web \cite{Sharma:etal:2018}. As discussed in more detail in the qualitative analysis below, when taken out of context, such captions are often non-informative, and thus it's not clear that learning to reproduce their style as closely as possible, like vanilla ClipCap-ConCap does, is a good idea. Consider for example Fig.~\ref{fig:caption-examples}(c) below. ClipCap-ConCap is perfectly reproducing the ground-truth description (``digital art selected for the \#''), and yet this is not as informative as the caption produced by DiscriTune-ConCap (``a boy in a pond with a lot of stars''). We thus conjecture that, despite their lower faithfulness to the human ground truth, DiscriTune-ConCap captions are more informative than ClipCap captions, and possibly even more informative than the original human captions in this dataset.

To verify this hypothesis, we designed an experiment in which human annotators had to select a target image from a set of 10 candidates, based either on a ground-truth human description, or a ClipCap-ConCap or DiscriTune-ConCap-generated one. We sampled the data for this experiment from our Conceptual Captions test set. To make sure that the captions needed to be genuinely informative in order to allow successful target retrieval, we selected hard distractors among the nearest visual neighbours of the target. The latter were found by passing images through the CLIP ViT-B/32 visual encoder and computing the cosine of the resulting representations.\footnote{We found that very high-similarity neighbours are often near duplicates of the target image and we excluded those with similarity above $0.8$} We collected  human data for 500 target+distractor sets, for each of the three caption types. We recruited Amazon Mechanical Turk participants who saw 100 sets each.\footnote{\url{ https://www.mturk.com/}} Experimental details (including ethical approval) are in Appendix A.

\begin{table}[tb]
    \centering
    \begin{tabular}{l|r}
    \textit{Captions} & accuracy\\
    \hline
    Human &42.8\\
    ClipCap-ConCap &36.2\\
    DiscriTune-ConCap &\textbf{47.6}\\
    \end{tabular}
    \caption{Percentage accuracy of human annotators when selecting a target image among 9 distractors based on the target human-, ClipCap- or DiscriTune-generated captions.}
    \label{tab:human-discrimination}
\end{table}

Results are reported in Table \ref{tab:human-discrimination}. They confirm that the task, due to the hard distractors, is quite challenging, with the annotators failing to reach 50\% accuracy independently of caption type. However, even with the least informative ClipCap-ConCap captions, humans are well above the 10\% chance level. Importantly, the DiscriTune-ConCap captions greatly outperform not only their ClipCap counterparts, but also the human ones, with a solid 5\% accuracy boost. We thus confirm our conjecture that, given a relatively noisy dataset such as Conceptual Captions, our finetuning procedure can produce captions that are more informative than the original human-produced descriptions. In the next section, we explore the differences between human and DiscriTune-generated captions.

\section{Caption Analysis}
\label{sec:caption-analysis}

\begin{figure*}[t]
    \centering
    \includegraphics[width = \hsize]{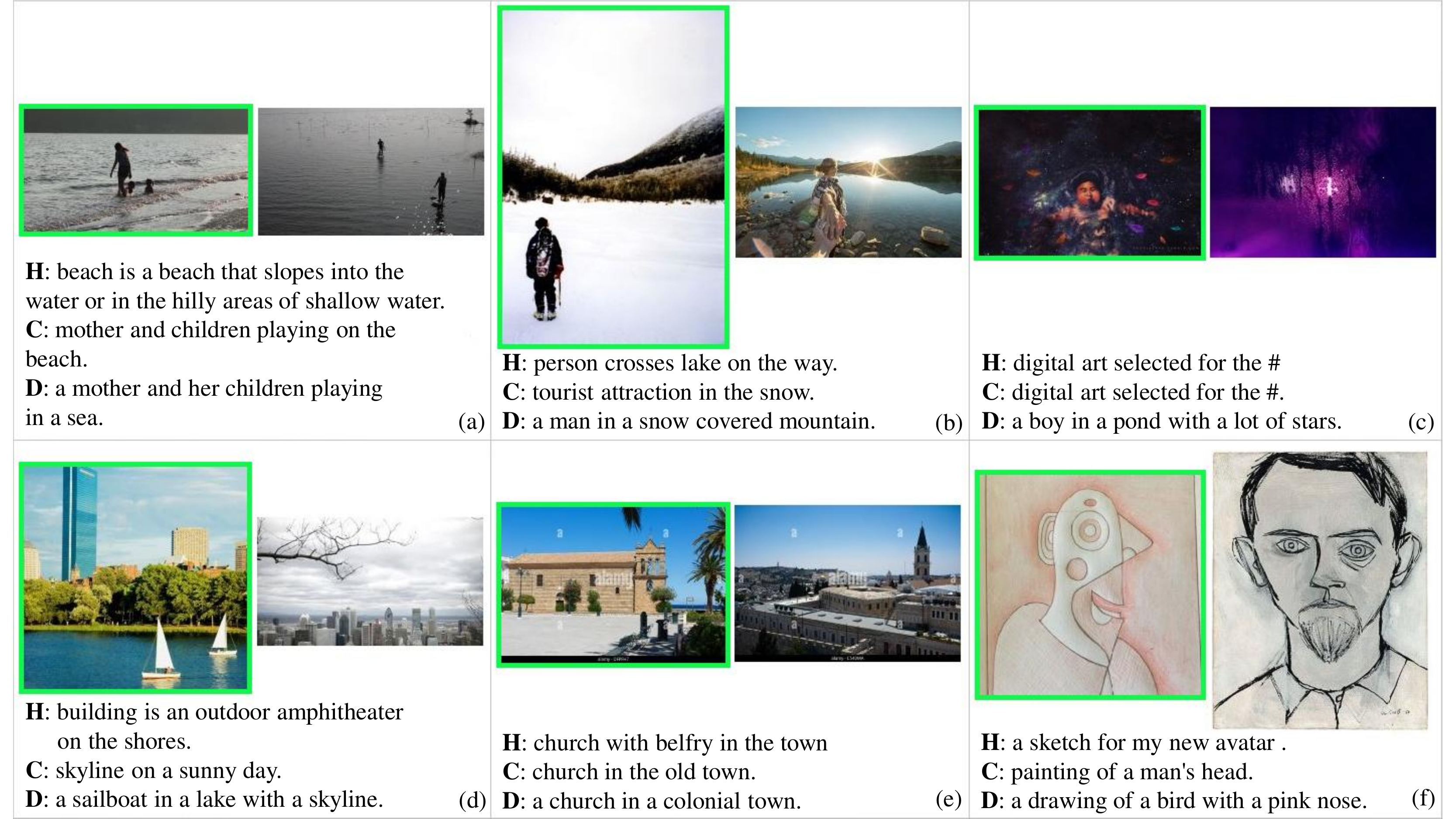}
    \caption{Example Conceptual Captions images with the corresponding captions produced by humans (\textbf{H}), ClipCap (\textbf{C}) and DiscriTune (\textbf{D}), respectively. In all cases, annotators were able to identify the target image (framed in green) only when receiving the DiscriTune caption as input. The image on the right of each pair is the distractor that was chosen by annotators when receiving the human caption as input.}
    \label{fig:caption-examples}
\end{figure*}

Fig.~\ref{fig:caption-examples} shows 6 Conceptual Captions examples from the experiment with humans in which human annotators were only able to guess the target given the DiscriTune-generated captions (correct target image is on the left; image picked by subjects in response to the human-generated caption is shown on the right). These cases, selected among those where no human face is recognizable, were hand-picked to illustrate various phenomena. They are however representative of the data, which we manually inspected as a whole.

Starting with a comparison of DiscriTune and ClipCap captions,\footnote{As the focus of this analysis in on the Conceptual Captions-trained/finetuned models, we will drop the -ConCap suffix.} we observe that the latter tend to be more vague (cf.~examples (b), (c), (d) and (f)), where sometimes these vague captions are actually matching the human-generated ones very closely (as in (c)). The other issue with ClipCap captions is that they are often inaccurate, as in example (a), where people are playing in the water, not on the beach; and it's hard to believe the desolated landscape in (b) is a tourist attraction. Note that the use of a generic term such as \textit{tourist attraction} in this last example is a peculiarity that ClipCap inherited from the Conceptual Captions style (the dataset was constructed by replacing place and people names with generic hypernyms such as this one). Clearly, learning to reproduce such Conceptual Captions-specific idiosyncrasies penalizes ClipCap when it's evaluated on other datasets. On the other hand, it's remarkable that discriminative finetuning, built on top of this very ClipCap system, was able to steer the captions back towards a more descriptive and precise language.

Comparing DiscriTune with the human captions, we see that the former tend to be more plainly descriptive and precise than the latter. Often, human captions in the Conceptual Captions dataset contain non-discriminative ``meta'' information about an image that is not useful to identify it, or might have made sense in the context of the original web page, but becomes opaque once the image and its alt-text are extrapolated. In example (a), the human caption is just stating that we are on a beach with shallow water, so that the human discriminator picked another shallow-beach picture. The DiscriTune caption precisely reports that there are a woman and children in the water. The human caption of example (b) might be accurate, but without more context it's difficult to recognize the white plateau as a frozen lake. Consequently, the human discriminator wrongly chose a more stereotypical picture of a lake. Concerning examples (c) and (f), the human captions report that they are ``digital art'' and an ''avatar'', respectively, leading the annotators to pick other potential exemplars of digital art and avatars from the distractors. The DiscriTune caption for (f) is actually inaccurate, as the avatar is not a bird, but the mention of a pink nose nevertheless helped the subject identifying the right image. The human caption of example (d) refers to an outdoor amphitheater that is not visible in the picture. Again, DiscriTune was more helpful to the human discriminators by providing a plain description of the picture contents. Finally, example (e) is interesting because the human caption highlights the somewhat atypical belfry in the picture (and so the human discriminator picked a photo with a more prominent belfry), whereas the DiscriTune caption provides a more discriminative cue by mentioning the colonial style of the landscape.

\begin{table}[tb]
    \centering
    \begin{tabular}{c|c|c}
    \textit{human}&\textit{ClipCap}&\textit{DiscriTune}\\
    \hline
    new&	young&    red\\
    other&old&    green\\
    
    musical&biological&blue\\
    outdoor& close& 	 black\\
    
    small&	white&       pink\\
    long&	trendy& 	 colorful\\
    
    big& digital&    yellow	 \\
    low& funny&	 denim\\
    
    large&aerial&      purple	\\
    beautiful&     general&     	 high\\
    \end{tabular}
    \caption{Top 10 adjective lemmas most associated to a caption type (\textit{human}, \textit{ClipCap} or \textit{DiscriTune}) according to the local Mutual Information association statistics computed on all captions generated for our full Conceptual Captions test set.}
    \label{tab:typical_adj}
\end{table}

To get a more general sense of the language of the different caption types, we lemmatized and part-of-speech tagged the full Conceptual Captions test set caption corpora with Spacy.\footnote{\url{https://spacy.io/}} We then computed the local Mutual Information statistics \cite{Evert:2005} across all possible lemma/caption-type pairs. We restricted the analysis to adjectives and nouns, as we found these two parts of speech to include the most visually descriptive terms. The adjective results are presented in Table \ref{tab:typical_adj} (noun analysis is in Appendix E). The difference between the adjectives in human and DiscriTune captions is striking: the latter are nearly all highly visual descriptive terms (in particular, colours), whereas the former contain several terms that are hardly providing any visual information (\textit{new}, \textit{musical}, \textit{other}). Even the most concrete human-caption adjectives are not as specific as those strongly associated with DiscriTune (compare \textit{small, big, large, beautiful} to \textit{red, blue, purple, denim}). The ClipCap list also contains few adjectives that might be genuinely useful to discriminate a specific image (\textit{white}, \textit{aerial}), with most being either abstract or very generic (\textit{biological}, \textit{trendy}, \textit{funny}, \textit{general}). 
It is remarkable that self-supervised discriminative finetuning, probably by exploiting the multimodal knowledge encoded in the pre-trained ClipCap components, is able to recover a highly visually descriptive language, despite the fact that it operates on a system that has been trained on human captions that, as we have just seen, are not as plainly descriptive, and that the system is not exposed to any new language during finetuning.

\section{Conclusion}

We presented a simple finetuning method to make model-generated captions more discriminative. Given a pre-trained captioner, its text generation component is finetuned on the task of helping a black-box text-based image retriever picking a target image among distractors. The task only requires unannotated images, and we were able to make the system work with the basic REINFORCE algorithm. 
We leave the exploration of more sophisticated reinforcement learning techniques as an obvious direction for future work. 
Our results are reported using two captioners, ClipCap, a decoder-only model, and BLIP, an encoder-decoder trained with multitask learning on web-scale data.

We found that the discriminatively finetuned captions do not improve over the original ones in terms of similarity to human ground truth, when tested on the same dataset the captioner was trained on. However, for both models and on a variety of out-of-domain datasets, they consistently outperform those of the original captioner. Discriminative pressure might be a strong enough signal to “unlearn” some of the overfit on the caption style of the pre-training dataset, and instead better capture the semantic content of the image. What's more, for the noisily annotated Conceptual Captions data-set (where we observed the largest performance drop in terms of mimicking ground-truth descriptions when finetuning ClipCap), discriminatively finetuned captions are more helpful than ground-truth captions, not only to a neural retriever, but also for humans tasked with a challenging image identification task. This suggests that our system could be used as-is to generate Web image captions that are on average more informative to users than alt-text descriptions (which are the source of Conceptual Captions annotations).

Qualitatively, we find that, even when finetuning the Conceptual Captions-trained captioner (that has learned to reproduce the somewhat abstract style of alt-text descriptions), our discriminative finetuning procedure recovers a more precise and plainly descriptive language. Compared to those of the original captioner, it is also clear why these more descriptive captions, that shed the idiosyncrasies of alt-text, will generalize better to other datasets.
We focused our analysis on ClipCap and Conceptual Captions, since this is the setup where we observed the largest discrepancy between human and discriminatively-generated captions, and a marked asymmetry in retrieval vs.~generation performance. We leave a thorough investigation of how the nature of the captions used to train the backbone captioner affects our method to future work. 
Given that several pre-trained text-based image retrievers are publicly available, another interesting direction would be to alternate different retrievers during finetuning. This might help the model further generalize, as it would be less likely to overfit the quirks of one specific retriever.

Finally, there has been recent progress in training models to learn from human feedback through reinforcement learning~\cite{Ouyang:etal:2022,Stiennon:etal:2020,Nakano:etal:2021}. Given the costly human annotation process required by this approach, our method could be seen as a cheaper alternative, exploiting ``neural'' feedback to guide the finetuning of an existing model. Future research directions should study the interplay between human and neural feedback to improve the capabilities of current systems.

\section*{Acknowledgements}
We thank the area chairs and reviewers for feedback. We thank Corentin Kervadec for feedback and help in creating Figure \ref{fig:setup}, and Lorenzo Baraldi and Manuele Barraco for replying to a Sunday night email during the rebuttal. EG, NCR, FF and MB received funding from the European Research Council (ERC) under the European Union's Horizon 2020 research and innovation programme (grant agreements No.\ 715154 and No.\ 101019291) and the Spanish Research Agency (ref.\ PID2020-112602GB-I00/MICIN/AEI/10.13039/501100011033). This paper reflects the authors' view only, and the funding agencies are not responsible for any use that may be made of the information it contains.

{\small
\bibliographystyle{ieee_fullname}
\bibliography{marco,others}
}

\appendix
\section{Crowdsourcing Experiment Details}
\label{appendix:crowdsourcing}

We chose the stimuli for the human annotation experiment as follows. We iterated over the images in our Conceptual Captions test set, and sampled, for each image, the 9 closest neighbours, thus creating sets composed of 10 images: 1 target and 9 distractors. We set a threshold of maximum cosine similarity to the nearest neighbour of $0.8$. We decided on this threshold after visual inspection of the generated sets: we aimed at having sets challenging for the annotators, yet not impossible to solve, and higher thresholds could lead to sets containing almost identical images, such as subsequent frames extracted from the same video or different croppings of the same picture. Neither targets nor distractors were repeated in the sets and we manually excluded disturbing images.

\begin{figure}[tb]
    \centering
    \includegraphics[width=\linewidth]{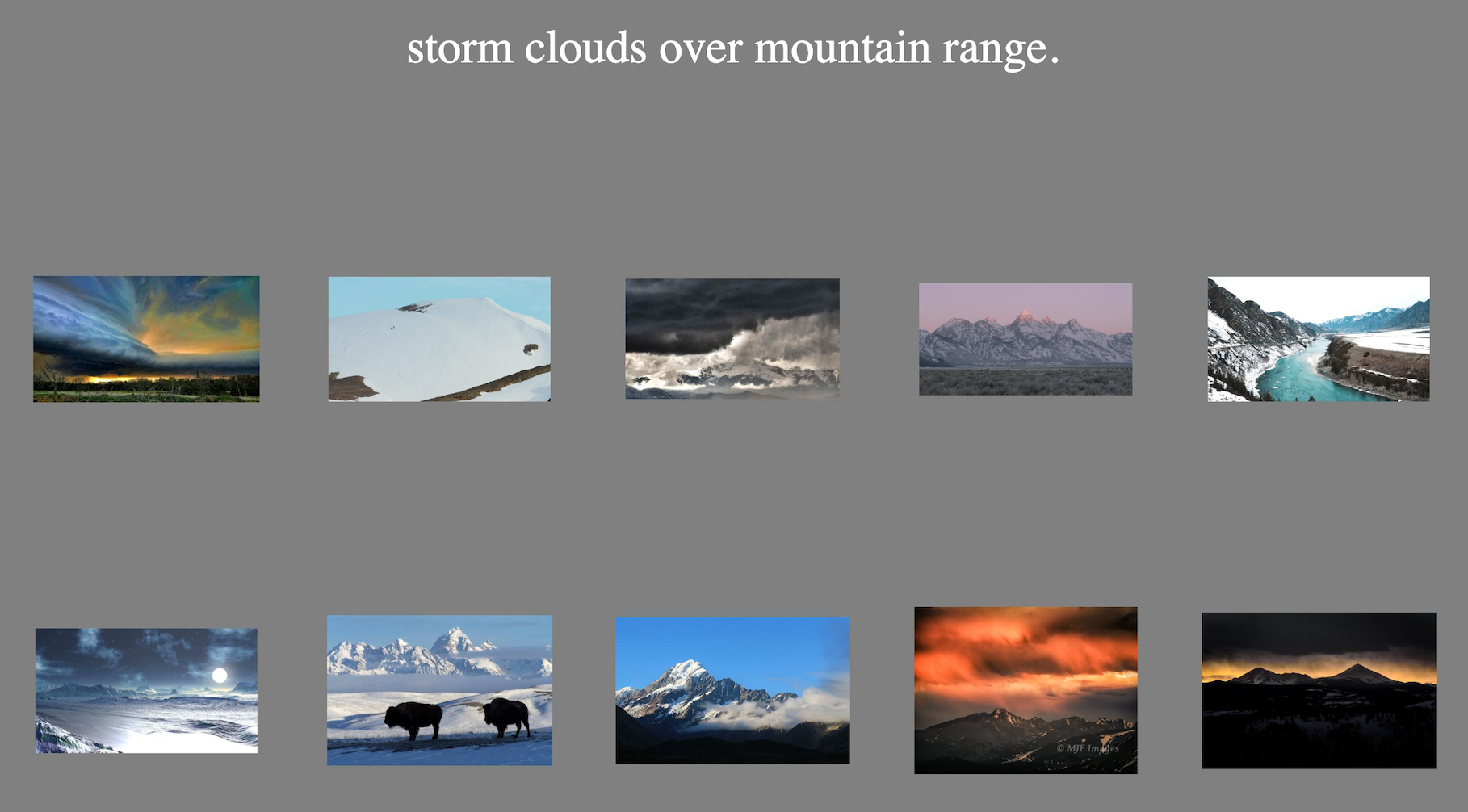}
    \caption{Example of a screen shown to the participants, with a human caption.}
    \label{fig:pavlovia}
\end{figure}

In the human retrieval experiment, we annotated each set with 3 types of captions: human captions, or captions generated by DiscriTune(-ConCap) or ClipCap(-ConCap), respectively. We randomly divided the entire set into blocks of 100 questions containing mixed caption types. On each screen, the 10 images from a set were presented at the center, arranged in two arrays of 5 images, with the caption written above--see Figure \ref{fig:pavlovia}.
Participants were asked to click on the image that matched the caption best. They were shown one example before starting the task, and were also warned that some cases could contain automatically-generated captions: we asked them to always reply with the answer they found most plausible. Finally, they were warned that the experiment contained some control items, used to ensure annotation quality.

Each subject was presented with one block of questions, plus 5 randomly placed controls, designed to ensure that annotators were paying attention to the task. These cases were made intentionally very simple: targets were surrounded by random distractors not appearing in the other sets, and the associated caption was a human generated one. We made sure internally that these sets could be easily processed with 100\% retrieval accuracy.

The data collection routine was written in Psychopy \cite{Peirce2019} and launched through Pavlovia.\footnote{\url{https://pavlovia.org/}} There was no time limit for completing the study. We recruited participants via Amazon Mechanical Turk.\footnote{\url{https://www.mturk.com}} We only accepted annotators from the US, with HIT approval rate higher than 97\%, and number of approved HITs higher than 1000. We informed them that we would not collect any personal data (except for their workerID, necessary for their payment, that we would not make public), and that the goal of the experiment was to study how well people identify images based on descriptions. Before being able to access the link of the experiment, participants had to complete an informed consent form, warning them that our experiment would show images and descriptions sampled from the web, and that could therefore contain upsetting content (although, as said above, we manually ensured that images we personally found disturbing would be excluded). They were able to quit the experiment at any time. We paid them $13.5$\$ for completing the task. The experiment was approved by the ethical board of Universitat Pompeu Fabra in the context of the AMORE project (grant agreement No.\ 715154). We excluded the data of participants that made more than one mistake when scoring the controls, suggesting that they were not paying enough attention to the task.

\section{Hyperparameter Exploration}
\label{appendix:ablation}
To generate text, we use greedy decoding at train time and beam size with 5 beams at test time, without tuning these values.
hyperparameter searches, we use retrieval score as our selection metric, since its consistent with our training objective and does not require annotated data.
We perform our hyperparameter searches on the Flickr validation set. Even though finetuning on Flickr might be seen as favouring out-of-domain performance (the models were pre-trained on COCO or Conceptual Captions), we we are confident it is not a major factor: the following sections show that the type of REINFORCE baseline and reward function do not have a big impact on performance, and for learning rate we informally found that, as long as large values are avoided, it does not greatly affect results, only convergence speed.
Finally, optimizing text-based models with reinforcement learning, especially when done form scratch, can be a challenging problem due to sparse rewards and the vast action space of selecting a token from a vocabulary. This can lead to repetitions or other unnatural word sequences. In practice, we did not notice issues such as repetitions or ungrammatical text. Indeed, we observed quite the opposite, with DiscriTune consistently producing natural text, as confirmed by the NLG metrics improvements. We believe this is due to starting from pre-trained models that are producing fluent language. Evidently, discriminative REINFORCE tuning does not degrade fluency.

\subsection{Reinforce Baseline}
Using REINFORCE~\cite{Williams:1992}, we can rewrite the gradient of the expected reward as the expectation of the gradient, approximated by a single sample caption $\hat c$: 
\begin{equation}
\begin{split}
\nabla_{\theta} L(i, D, \theta)
&= \nabla_{\theta} \mathbb{E}_{c \sim P_\theta(\cdot|i)}[-R(c,i,D)] \\
&= \mathbb{E}_{c \sim P_\theta(\cdot|i)}[-R(c,i,D) \nabla_{\theta} \log P_\theta(c|i)] \\
&\approx -R(\hat c, i, D) \nabla_{\theta} \log P_\theta(\hat c|i)
\end{split}
\end{equation}

\noindent{}where $i$ is the target image, $\hat c$ is the generated caption, $D$ is the set candidates fed to the retriever and $R$ is the reward function.
The parameters $\theta$ can be optimized with regular (mini-batch) gradient descent.
To reduce variance of the gradient estimator when using REINFORCE, we subtract a baseline term. We compared two different baselines. The first is a running mean of past rewards values using greedy decoding. The second uses beam with the baseline computed using the reward value given by CLIP when fed captions generated with greedy decoding. We trained a ClipCap model on Flickr for 10 epochs using the setup described in Section~4, and then evaluated its performance on the validation set. Results are presented in Table~\ref{tab:ablation-baseline}.
We found that, without subtracting a baseline, the model performed poorly, achieving an accuracy of 34.3\%. Running mean and greedy decoding yield similar performance, with running mean showing slightly higher accuracy (97.8\% vs 97.4\%). We thus employed a running mean baseline with greedy decoding in all the main experiments.

\begin{table}[H]
    \centering
    \begin{tabular}{lc}
    baseline type & P@1 \\
    \hline
    no baseline & 34.3 \\
    greedy decoding (w/ beam search) & 97.4 \\
    running mean & 97.8 \\
    \hline\hline
    \end{tabular}
    \caption{ClipCap retrieval accuracy with 100 candidates on the Flickr validation set using different REINFORCE baselines. The \textit{no baseline} and \textit{running mean} methods were used employing greedy decoding to generate captions, whereas when \textit{greedy decoding} was the baseline, we let ClipCap produce captions with beam search using 5 beams.}
    \label{tab:ablation-baseline}
\end{table}

\subsection{Reward Function}
\label{app:reward-hp}

To find the best reward to train our captioner, we trained a ClipCap model on Flickr for 10 epochs using the setup described in Section~4, and then evaluated its performance on the validation set. We explored three different reward functions. The cosine similarity reward computes the normalized dot product between the target image embedding and the model-generated caption representation. This is equivalent to the CLIPScore \cite{Hessel:etal:2021} and it is not discriminative since it does not compare the target image with any distractor.
The accuracy-based reward computes a binary score which is 1 if CLIP assigned the highest dot-product-based alignment score to the target image when fed a caption, and 0 otherwise. The third reward type is the negative softmax-normalized log probability of the match between a caption and each image in the candidate list, as described in Section~3.
As reported in Table \ref{tab:ablation-reward}, the log probability reward performed best, although not by a large margin. Thus, we run all the experiments presented in this work optimizing the captioner using such reward.

\begin{table}[H]
    \centering
    \begin{tabular}{lc}
    reward function & P@1 \\
    \hline
    cosine similarity & 85.2 \\
    accuracy & 85.3 \\
    log probability & 86.2\\
    \hline\hline
    \end{tabular}
    \caption{ClipCap retrieval accuracy with 100 candidates on the Flickr validation set with different reward functions.}
    \label{tab:ablation-reward}
\end{table}

\section{Finetuning CaMEL}
We apply our DiscriTune method to the recently introduced CaMEL \cite{Barraco:etal:2022} captioner model. This model is trained on COCO using a distillation loss based on a model tracking the running mean of an online network, and concurrently optimized with a reward-based objective after a first phase of supervised learning against human references. The reward is computed using CIDEr~\cite{Vedantam:etal:2015} (please see  \cite{Barraco:etal:2022} for additional details on the model and its training setup).
NLG results in Table~\ref{tab:camel-results-nlg} confirm that, at the price of a small drop in in-domain performance, DiscriTune is able to improve (by a small margin) when tested on the Flickr out-of-domain dataset, confirming the benefits of discriminatively finetuning a pre-trained captioner, even when the procedure is applied to a ``bleeding-edge'' model of this sort.

\begin{table}[tb]
    \centering
    \begin{tabular}{l|cccc}
    \multicolumn{5}{c}{\textit{COCO}}\\ 
    \hline
    Model& B@4& M& C& S \\
    CaMEL&  \textbf{38.11}& \textbf{29.03} & \textbf{128.62} & \textbf{23.35}\\
    DiscriTune-CaMEL& 33.45& 27.63& 117.71& 22.03 \\
    \hline\hline
    \multicolumn{5}{c}{\textit{Flickr}}\\
    CaMEL & \textbf{22.93} & 20.93 & 58.38 & 14.62 \\
    DiscriTune-CaMEL & 22.60 & \textbf{20.99} & \textbf{59.12} & \textbf{14.94}\\
    \hline\hline 
    \end{tabular}
    \caption{NLG metrics (BLEU@4\cite{Papineni:etal:2002}, METEOR\cite{Denkowski:etal:2014}, CIDEr\cite{Vedantam:etal:2015} and SPICE\cite{Anderson:etal:2016}) for CaMEL and DiscriTune-CaMEL captions on the COCO test split (in-domain, our results when using CaMEL) and Flickr test split (out-of-domain).}
    \label{tab:camel-results-nlg}
\end{table}

\section{Image Retrieval with Hard Distractors}
\label{sec:hard-distractors}

\subsection{ImageCoDe}
In order to test retrieval performance in a challenging setup, we use the \textbf{ImageCoDe} dataset \cite{Krojer:etal:2022}. ImageCoDe was recently introduced as a testbed for text-based image retrieval. It is formed by $10$-elements sets of target images collected from consecutive video frames or by mining similar images to a given target frame. For a fair comparison with prior work, we use the validation images as test data.

In Table~\ref{tab:imagecode}, we report ImageCoDe results with all our model-generated captions as well as human-generated ones, when a CLIP model with ViT-B-32 was used as the text-to-image retriever. The models are the ones trained with the setup described in Section 4. The results are remarkable, reaching a new state of the art (for either human or model-generated captions) on this dataset (best previous result, obtained with human captions: 29.9\% \cite{Krojer:etal:2022}).  This shows that our tuning method is beneficial to produce discriminative captions even in contexts in which distinctions need to be very subtle. This suggests that our method could be profitably applied to scenarios where such granular discrimination is called for, such as in video understanding tasks.

\begin{table}[tb]
    \centering
    \begin{tabular}{l|r}
    \textit{Captions} & ImageCoDe\\
    \hline
    ClipCap-COCO &28.7 \\
    DiscriTune-COCO &34.0 \\
    \hline
    ClipCap-ConCap & 26.8 \\
    DiscriTune-ConCap & \textbf{36.2} \\
    \hline
    Blip-COCO & 24.0 \\
    DiscriTune-COCO & 24.7 \\
    \hline
    Human & 22.3\\
    \hline\hline
    \end{tabular}
    \caption{Percentage accuracy (P@1) when retrieving a target image
    from the validation image sets of ImageCoDe. Random chance is at 10\%.}
    \label{tab:imagecode}
\end{table}

\subsection{Hard Negative Mining}
We perform an additional experiment where at training time the retrieval task is performed using  automatically-mined hard distractors. When testing, we still randomly select all non-target candidates. We pick the \textit{k} most similar distractors based on the cosine similarity with a target using the CLIP visual encoder (the remaining $99-k$ distractors are picked randomly, as usual). This experiment is aimed at studying the impact of the distractors in discriminative finetuning, with the idea that making the task harder should lead to more discriminative captions.
In Table~\ref{tab:gen-results-nlg-hard-negatives} and Table~\ref{tab:ret-results-nlg-hard-negatives} we report NLG metrics and retrieval accuracy, respectively. Overall, we see a mixed picture. With respect to the NLG metrics, hard distractors are helpful only for one of the two OOD datasets (Flickr), but at the price of a larger performance drop in-domain (COCO). Concerning retrieval accuracy, hard distractors give a slight improvement over random ones for in-domain data (COCO) and only on Flickr but not on Conceptual Captions for out-of-domain data.

We conjecture that the harder setup can lead to overfitting the quirks of the frozen retriever, in some cases leading to (slightly) poorer generalization. Studying the impact of the retrieval task with respect to number and type of distractors is an interesting direction for future work.

\begin{table}[tb]
    \centering
    \begin{tabular}{l|cccc}
    \multicolumn{5}{c}{\textit{COCO}}\\ 
    \hline
    Model& B@4& M& C& S\\
    ClipCap-COCO&  \textbf{32.60}& \textbf{27.50}& \textbf{108.55}& \textbf{20.33}\\
    DiscriTune-COCO& 32.31& 26.05& 105.40& 20.03 \\
     w/ 5 hard distractors & 29.85& 25.53& 100.25& 19.50\\
     w/ 10 hard distractors & 29.20& 25.25& 98.69& 19.30 \\
    \hline\hline
    \multicolumn{5}{c}{\textit{Conceptual Captions}}\\ 
    \hline
    ClipCap-COCO&  1.47& 6.43& 23.74 & 7.98 \\
    DiscriTune-COCO& \textbf{1.71}& \textbf{6.58}& 
    \textbf{28.01}& \textbf{9.00}\\
    w/ 5 hard distractors & 1.47 & 6.22 & 25.11 & 8.50\\
    w/ 10 hard distractors & 1.39 & 6.15 & 24.69 & 8.40\\
    \hline\hline
    \multicolumn{5}{c}{\textit{Flickr}}\\ 
    \hline
    ClipCap-COCO& 17.21& 18.43& 41.65& 12.04\\
    DiscriTune-COCO& 18.48& 18.61& 44.78& 12.68\\
    w/ 5 hard distractors & \textbf{18.75} & \textbf{18.95} & \textbf{45.15} & \textbf{13.00}\\
    w/ 10 hard distractors & 18.23 & 18.68 & 44.28 & 12.85\\
    \hline\hline
    \end{tabular}
    \caption{NLG metrics (BLEU@4, METEOR, CIDEr and SPICE) for ClipCap and ClipCap-based DiscriTune captions on COCO, Conceptual\-Captions and Flickr, after training with 5 or 10 automatically mined hard distractors and testing with randomly selected ones.}
    \label{tab:gen-results-nlg-hard-negatives}
\end{table}

\begin{table}[tb]
    \centering
    \begin{tabular}{l|c}
    \textit{COCO} &\\ 
    \hline
    Model& P@1\\
    ClipCap-COCO&  74.2\\
    DiscriTune-COCO& 84.8\\
     w/ 5 hard distractors & 84.9\\
     w/ 10 hard distractors & \textbf{85.2} \\
    \hline\hline
    \textit{Conceptual Captions} & \\ 
    \hline
    ClipCap-COCO& 73.0\\
    DiscriTune-COCO& \textbf{83.6}\\
     w/ 5 hard distractors & 82.3\\
     w/ 10 hard distractors &  82.0\\
    \hline\hline
    \textit{Flickr} &\\ 
    \hline
    ClipCap-COCO& 65.9\\
    DiscriTune-COCO& 79.4\\
     w/ 5 hard distractors & \textbf{79.8}\\
     w/ 10 hard distractors &  79.1\\
    \hline\hline
    \end{tabular}
    \caption{P@1 retrieval accuracy  for ClipCap and ClipCap-based DiscriTune captions on COCO, Conceptual Captions and Flickr, after training with 5 or 10 automatically mined hard distractors and testing with randomly selected ones.}
    \label{tab:ret-results-nlg-hard-negatives}
\end{table}

\section{Caption Analysis: Nouns}
\label{app:analysis}

The patterns we encountered in the adjective analysis  presented in section~5 %
are confirmed by the noun lemma analysis. In Table~\ref{tab:typical_nouns}, we report the top 10 noun lemmas most strongly associated with the human, Clipcap and DiscriTune captions in the Conceptual Captions dataset.
DiscriTune favours words with strong and precise visual content, such as \textit{dress}, \textit{woman}, \textit{pair}, \textit{garden}, \textit{field} and \textit{lake}. Human captions tend to include more generic terms such as \textit{person} and \textit{background}, as well as several nouns that might be describing images at a more abstract level, that would probably not favour their precise identification (\textit{image, time, summer, style, part}). The preference for these more abstract terms is even more pronounced in ClipCap (\textit{view, actor, portrait, illustration, premiere, artist, vector, property}).

\begin{table}[h!]
    \centering
    \begin{tabular}{l|l|l}
    \textit{human}&\textit{ClipCap}&\textit{DiscriTune}\\
    \hline
    \multicolumn{3}{c}{\textbf{nouns}}\\
    \hline
person&    view&forest    \\
time&   illustration&	 night\\

day&	    actor&	   dress\\
water& premiere&	     garden\\

image&	     person&	  celebrity\\
summer& artist&	 field\\

instrument& portrait&	 woman\\

style&vector &lake\\

background&background&        pair\\
part&   property&    group\\

    \end{tabular}
    \caption{Top 10 noun lemmas most associated to a caption type (\textit{human}, \textit{ClipCap} or \textit{DiscriTune}) according to the local Mutual Information association statistics computed on all captions generated for our full Conceptual Captions test set.}
    \label{tab:typical_nouns}
\end{table}

\end{document}